\documentclass[10pt,twocolumn,letterpaper]{article}

\usepackage{wacv}
\usepackage{times}
\usepackage{epsfig}
\usepackage{graphicx}
\usepackage{amsmath}
\usepackage{amssymb}
\usepackage{color}

\usepackage{graphicx}
\usepackage[export]{adjustbox}

\usepackage{amsfonts}
\usepackage{booktabs}
\usepackage{siunitx}
\usepackage{pifont}

\usepackage{subfig}
\usepackage{diagbox}
\usepackage{multirow}

\usepackage[pagebackref=true,breaklinks=true,letterpaper=true,colorlinks,bookmarks=false]{hyperref}

\wacvfinalcopy 

\def\wacvPaperID{145} 

\ifwacvfinal\fi
\begin{document}




\title{Word-level Deep Sign Language Recognition from Video:\\ A New Large-scale Dataset and Methods Comparison}

\def\HD#1{{\color{red}{\bf [HD:} {\it{#1}}{\bf ]}}}
\def\DX#1{{\color{cyan}{\bf [DX:} {\it{#1}}{\bf ]}}}
\definecolor{applegreen}{rgb}{0.55, 0.71, 0.0}
\def\CR#1{{\color{applegreen}{\bf [CR:} {\it{#1}}{\bf ]}}}
\def\XY#1{{\color{blue}{\bf [XY:} {\it{#1}}{\bf ]}}}

\author{Dongxu Li , Cristian Rodriguez Opazo, Xin Yu, Hongdong Li\\
The Australian National University, Australian Centre for Robotic Vision (ACRV)\\
{\tt\small \{dongxu.li, cristian.rodriguez, xin.yu, hongdong.li\}@anu.edu.au}
}

\maketitle

\ifwacvfinal\thispagestyle{empty}\fi

\textbf{}\begin{abstract}

Vision-based sign language recognition aims at helping deaf people to communicate with others. 
However, most existing sign language datasets are limited to a small number of words. 
Due to the limited vocabulary size, models learned from those datasets cannot be applied in practice.
In this paper, we introduce a new large-scale Word-Level American Sign Language (WLASL) video dataset, containing more than 2000 words performed by over 100 signers. This dataset will be made publicly available to the research community. To our knowledge,it is by far the largest public ASL dataset to facilitate word-level sign recognition research. 

Based on this new large-scale dataset, we are able to experiment with several deep learning methods for word-level sign recognition and evaluate their performances in large scale scenarios. Specifically we implement and compare two different models,i.e., (i) holistic visual appearance based approach, and (ii) 2D human pose based approach. Both models are valuable baselines that will benefit the community for method benchmarking.  Moreover, we also propose a novel pose-based temporal graph convolution networks (Pose-TGCN) that model spatial and temporal dependencies in human pose trajectories simultaneously, which has further boosted the performance of the pose-based method.  Our results show that pose-based and appearance-based models achieve comparable performances up to $62.63\%$ at top-10 accuracy on 2,000 words/glosses, demonstrating the validity and challenges of our dataset. Our dataset and baseline deep models are available at \url{https://dxli94.github.io/WLASL/}.

\end{abstract}
\section{Introduction}

\begin{figure}
    \centering
    \includegraphics[width=0.48\textwidth]{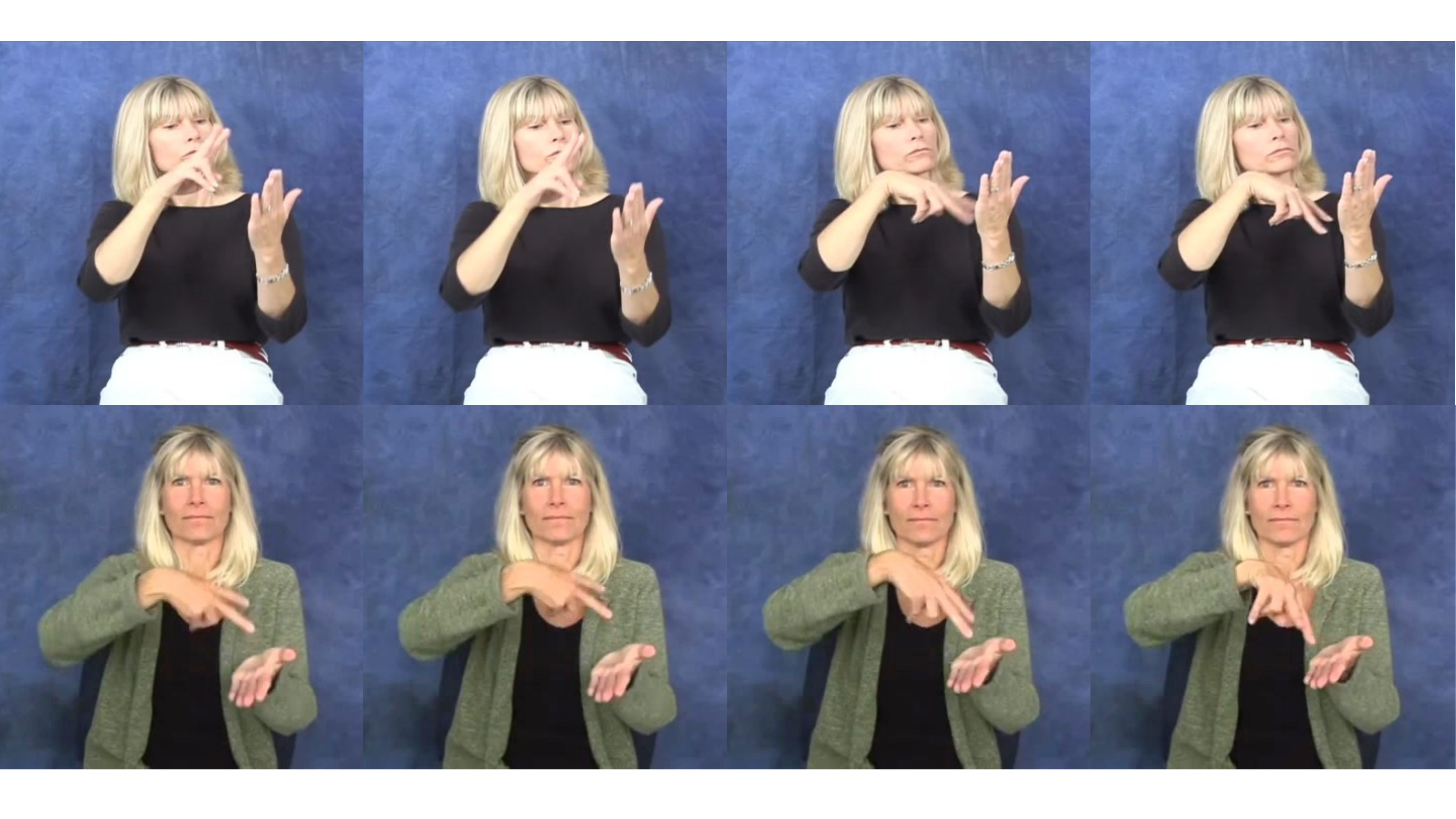}
    \vspace{-2em}
    \caption{ASL signs ``read" (top row) and ``dance" (bottom row)~\cite{caselli2017asl} differ only in the orientations of the hands.}
    \label{fig:read-dance}
    \vspace{-1em}
\end{figure}

Sign languages, as a primary communication tool for the deaf community, have their unique linguistic structures.
Sign language interpretation methods aim at automatically translating sign languages using, for example, vision techniques.
Such a process involves mainly two tasks, namely, word-level sign language recognition (or ``isolated sign language recognition") and sentence-level sign language recognition (or ``continuous sign language recognition"). In this paper, we target at word-level recognition task for American Sign Language (ASL) considering that it is widely adopted by deaf communities over 20 countries around the world~\cite{mccaskill2011hidden}.



\begin{figure}%
    \centering
    \subfloat[The verb ``\textbf{Wish}" (top) and the adjective ``\textbf{hungry}" (bottom) correspond to the same sign.
    \label{fig:hungry-wish}]{{\includegraphics[width=0.4\textwidth]{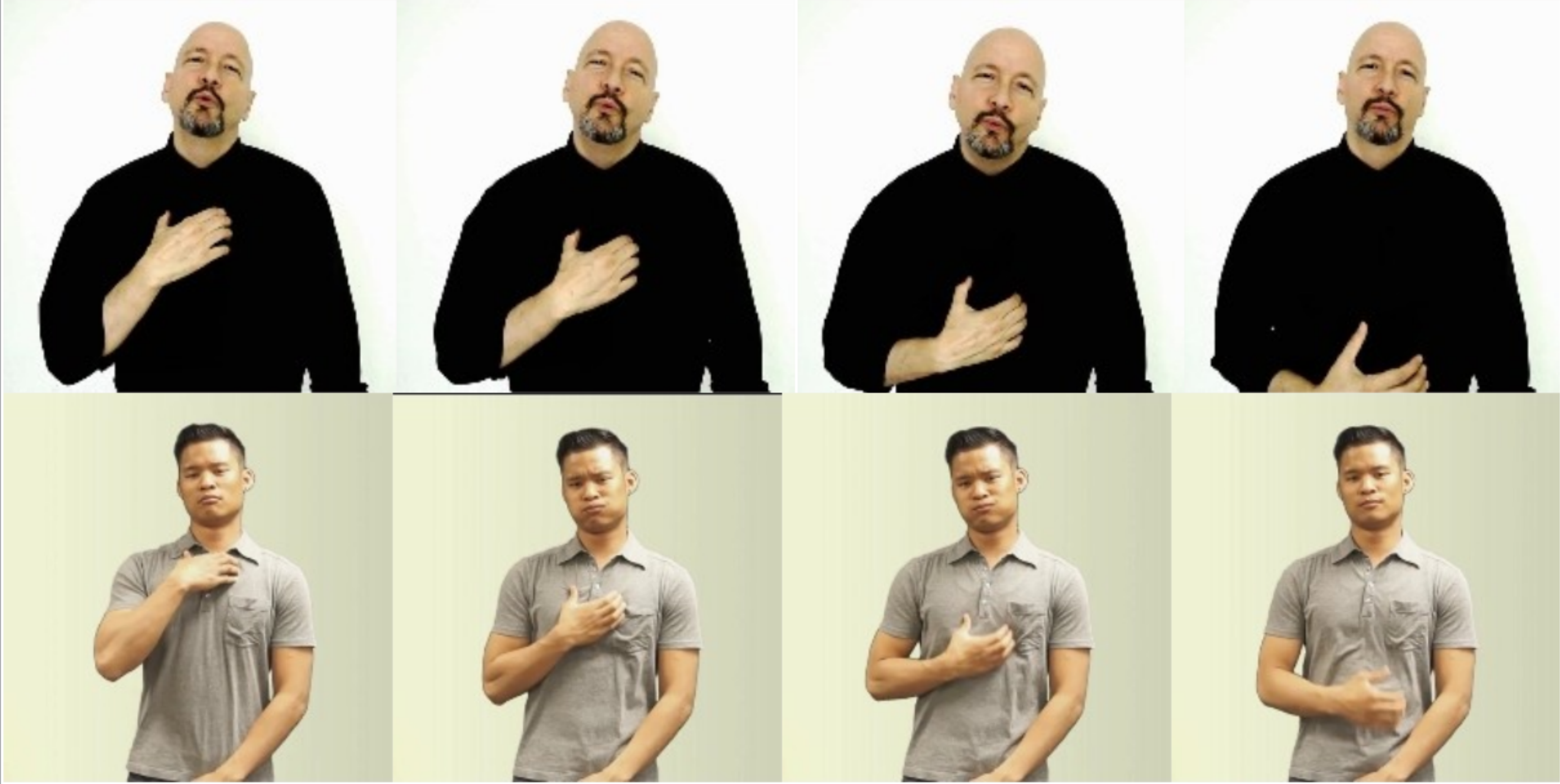} }}%
    \qquad
    \subfloat[The same sign represents different words ``\textbf{Rice}" (top) and ``\textbf{soup}" (bottom). ]{{\includegraphics[width=0.4\textwidth]{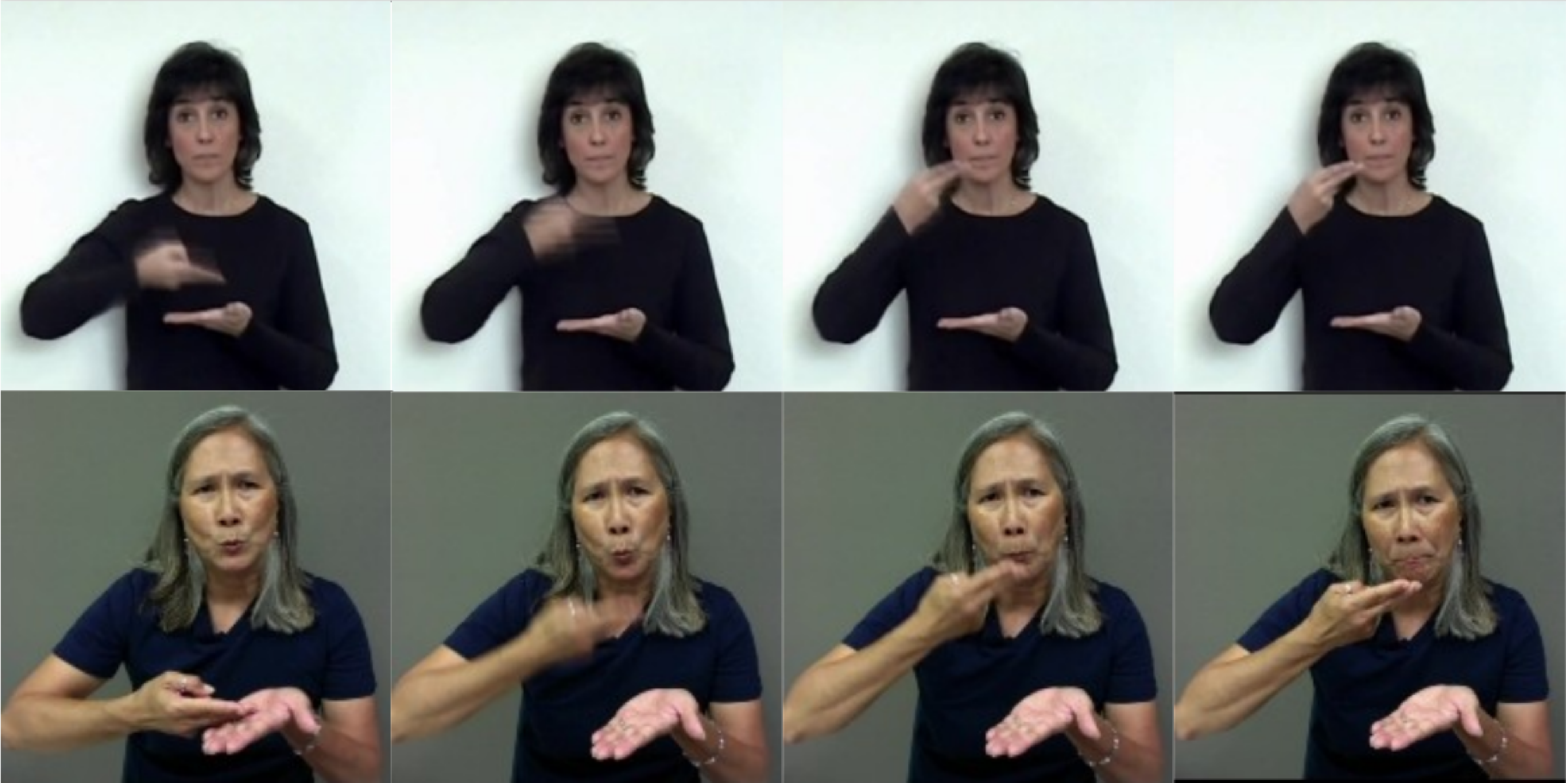} }}%
    \qquad
    \subfloat[Signers perform ``\textbf{Scream}" with different hand positions and amplitude of hand movements.\label{fig:scream}  ]{{\includegraphics[width=0.4\textwidth]{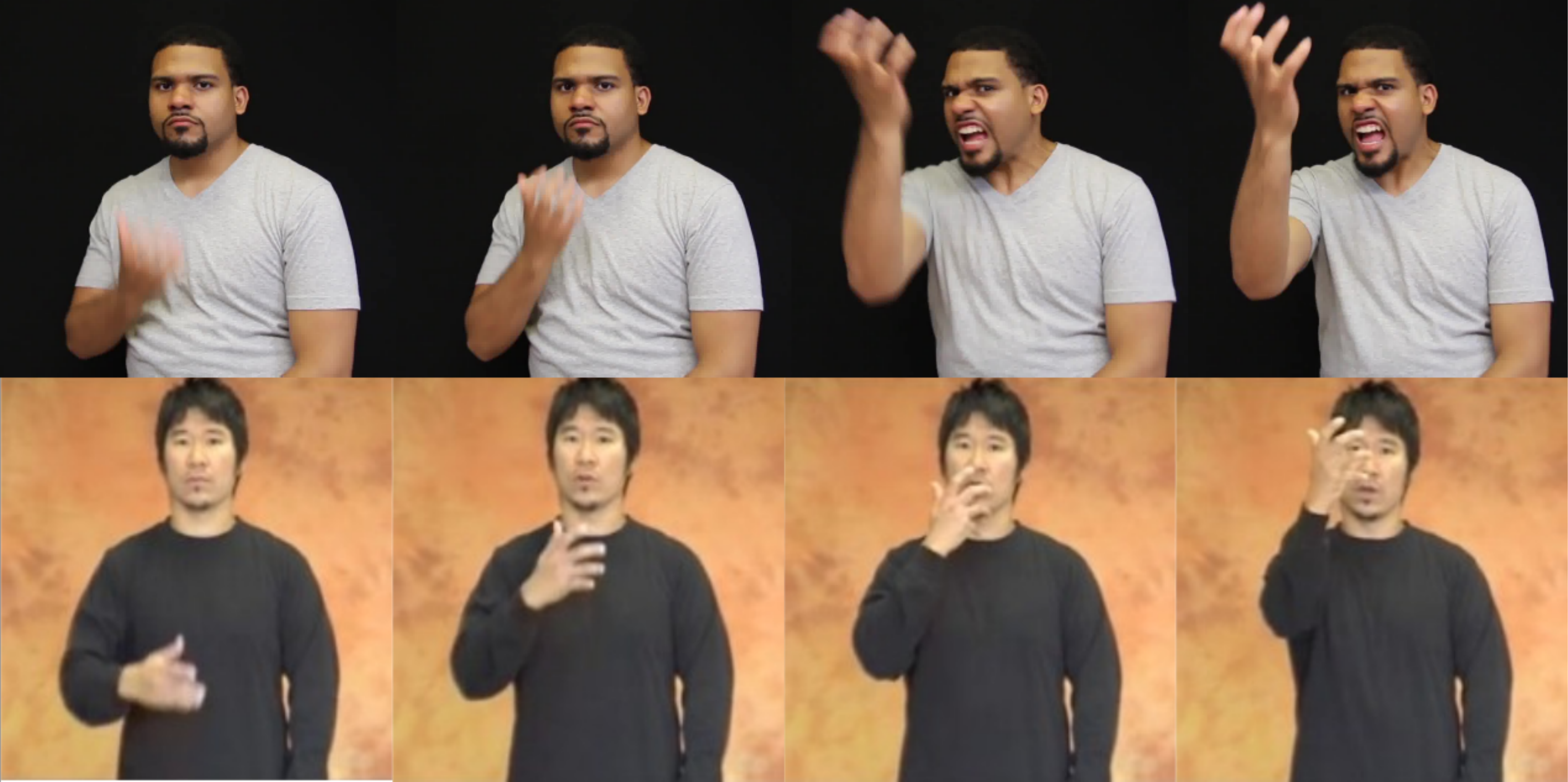} }}%
    \caption{Ambiguity and variations of Signing. (a, b) shows linguistic ambiguity in ASL. (c) shows signing variations of different signers. }%
    \vspace{-1em}
    \label{fig:rice-soup}%
\end{figure}

Serving as a fundamental building block for understanding sign language sentences, the word-level sign recognition task itself is also very challenging: 
\begin{itemize}
\vspace{-0.5em}
    \item The meaning of signs mainly depends on the combination of body motions, manual movements and head poses, and subtle differences may translate into different meanings. As shown in Fig.~\ref{fig:read-dance}, the signs for ``dance'' and ``read'' only differ in the orientations of hands.\vspace{-0.5em}
    \item The vocabulary of signs in daily use is large and usually in the magnitude of thousands. In contrast, related tasks such as gesture recognition~\cite{amir2017low,jester2019} and action recognition~\cite{jiang2014thumos,soomro2012ucf101,carreira2017quo} only contains at most a few hundred categories. This greatly challenges the scalability of recognition methods.
    \vspace{-0.5em}
    \item A word in sign language may have multiple counterparts in natural languages. For instance, the sign shown in Fig.~\ref{fig:rice-soup} (a), can be interpreted as ``wish'' or ``hungry'' depending on the context. In addition,
    nouns and verbs that are from the same lemma may have the same sign. These subtleties are not well captured in the existing small-scale datasets.
\end{itemize}
\vspace{-1mm}


In order to learn a practical ASL recognition model, the training data needs to contain a sufficient number of classes and training examples. Considering that existing word-level datasets do not provide a large-scale vocabulary of signs, we firstly collect large-scale word-level signs in ASL as well as their corresponding annotations. 
Furthermore, since we want to leverage the minimal hardware requirement for the sign recognition, only monocular RGB-based videos are collected from the Internet. By doing so, the trained sign recognition models do not rely on special equipment, such as depth cameras~\cite{kapuscinski2015recognition} and colored gloves~\cite{ronchetti2016lsa64}, and can be deployed in general cases. 
Moreover, when people communicate with each other, they usually sign in frontal views. Thus, we only collect videos with signers in near-frontal views to achieve a high-quality large-scale dataset. In addition, our dataset contains annotations for dialects that are commonly-used in ASL.
In total, our proposed WLASL dataset consists 21,083 videos performed by 119 signers, and each video only contains one sign in ASL. Each sign is performed by at least 3 different signers. Thus, inter-signer variations in our dataset facilitates the generalization ability of the trained sign recognition models. 

Based on WLASL, we are able to experiment with several deep learning methods for word-level sign recognition, based on (i) holistic visual appearance, and (ii) 2D human-pose.   
For \emph{appearance-based methods}, we provide a baseline by re-training VGG backbone~\cite{simonyan2014very} and GRU~\cite{cho2014learning} as a representative for convolutional recurrent networks. We also provide a 3D convolution networks baseline using fine-tuned I3D~\cite{carreira2017quo}, which performs better than the VGG-GRU baseline. 
For \emph{pose-based methods}, we firstly extract human poses from original videos and use them as input features. We provide a baseline using GRU to model the temporal movements of the poses. 
%
%
Giving that GRU captures explicitly only the temporal information in pose trajectories, it may not fully utilizes the spatial relationship between body keypoints.
Motivated by this, we propose a novel pose-based model \emph{temporal graph convolutional network (TGCN)} that captures the temporal and spatial dependencies in the pose trajectories simultaneously. Our results show that both pose-based approach and appearance-based approach achieve comparable classification performance on 2,000 words, reaching up to $62.63\%$.

\begin{figure*}[t]
    \centering
    \includegraphics[width=1\textwidth]{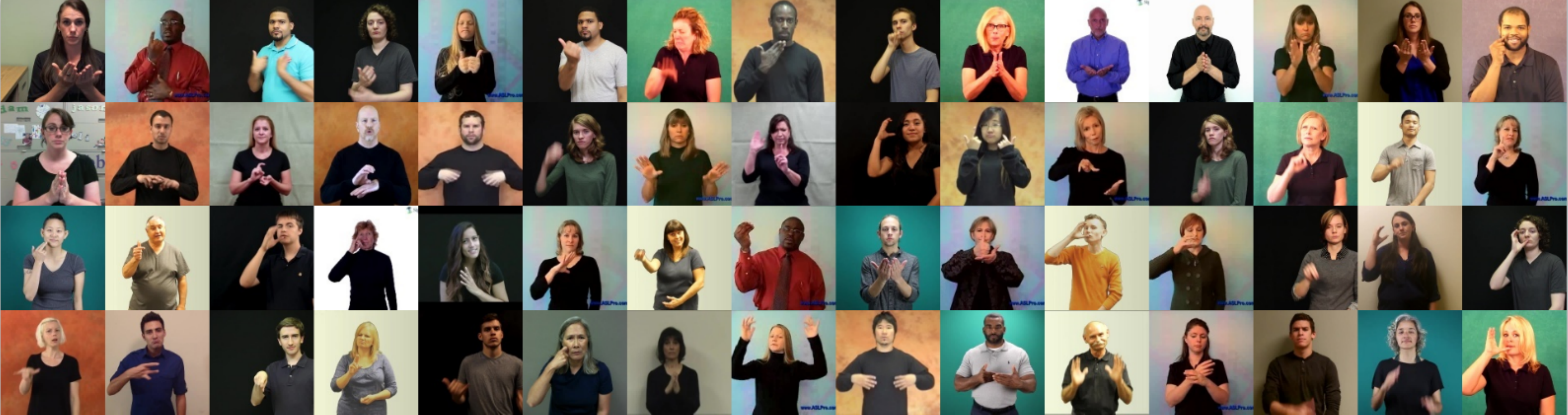}
    \caption{Illustrations of the diversity of our dataset, which contains different backgrounds, illumination conditions and signers with different appearances.}
    \label{fig:signer_profiles}
    \vspace{-1em}
\end{figure*}

\section{Related Work}\label{sec:related-work}

In this section, we briefly review some existing publicly sign language datasets, and state-of-the-art sign language recognition algorithms are also discuss to demonstrate the necessity of a large-scale ASL dataset.

\subsection{Sign Language Datasets}
There are three publicly released \emph{word-level ASL} datasets\footnote{We notice that one paper~\cite{joze2018ms} aims at providing an ASL dataset containing 1,000 glosses. Since the dataset is not released at the time of preparing the paper, we cannot evaluate and compare with the dataset.}, i.e. Purdue RVL-SLLL ASL Database~\cite{wilbur2006purdue}, Boston ASLLVD~\cite{athitsos2008american} and RWTH-BOSTON-50~\cite{zahedi2005combination}.

\textbf{Purdue RVL-SLLL ASL Database}~\cite{wilbur2006purdue} contains 39 motion primitives with different hand-shapes that are commonly encountered in ASL.  Each primitive is produced by 14 native signers. Note that, the primitives in \cite{wilbur2006purdue} are the elements constituting ASL signs but may not necessarily correspond to an English word. 
%
\textbf{Boston ASLLVD}~\cite{athitsos2008american} has 2,742 words (\ie, glosses) with 9,794 examples (3.6 examples per gloss on average).
Although the dataset has large coverage of the vocabulary, more than 2,000 glosses have at most three examples, which is unsuitable to train thousand-way classifiers.
%
\textbf{RWTH-BOSTON-50}~\cite{zahedi2005combination} contains 483 samples of 50 different glosses performed by 2 signers. Moreover,
\textbf{RWTH-BOSTON-104} provides 200 continuous sentences signed by 3 signers which in total cover 104 signs/words. 
\textbf{RWTH-BOSTON-400}, as a sentence-level corpus, consists of 843 sentences including around 400 signs, and those sentences are performed by 5 signers.
\textbf{DEVISIGN} is a large-scale word-level Chinese Sign Language dataset, consists of 2,000 words and 24,000 examples performed by 8 non-native signers in controlled lab environment. 
\begin{table}[t]\caption{Overview of word-level datasets in other languages.}
\vspace{-1em}
\centering
\resizebox{\linewidth}{!}{
\begin{tabular}{cccccc} \toprule
    \textbf{Datasets} & \textbf{\#Gloss} &\textbf{\#Videos} & \textbf{\#Signers} & \textbf{Type} & \textbf{Sign Language} \\ \midrule
    LSA64~\cite{ronchetti2016lsa64}  & 64 & 3,200    & 10 & RGB & Argentinian \\
    PSL Kinect 30~\cite{kingma2014adam}  & 30 & 300    & - & RGB, depth & Polish \\
    PSL ToF~\cite{kingma2014adam}  & 84 & 1,680    & - & RGB, depth & Polish \\
    DEVISIGN~\cite{chai2015devisign}  & 2,000 & 24,000    & 8 & RGB, depth & Chinese \\
    GSL~\cite{Efthimiou2007GSLCCA}  & 20 & 840    & 6 & RGB & Greek \\
    DGS Kinect~\cite{kinect40}  & 40 & 3,000    & 15 & RGB, depth & German \\
    LSE-sign~\cite{gutierrez2016lse}  & 2,400 & 2,400    & 2 & RGB  & Spanish \\
    \bottomrule
\end{tabular}
\vspace{-2em}
    \label{tab:non-eng}
}
\end{table}
Word-level sign language datasets exist for other regions, as summarized word-level sign language datasets in other languages in Table~\ref{tab:non-eng}. 

All the previously mentioned datasets have their own properties and provide different attempts to tackle the word-level sign recognition task. However, they fail to capture the difficulties of the task due to insufficient amount of instance and signer.

\subsection{Sign Language Recognition Approaches}\label{subsubsec:approaches}
Existing word-level sign recognition models are mainly trained and evaluated on either private~\cite{grobel1997isolated,kulkarni2010appearance,zafrulla2011american,huang2015sign,Pigou_2017_ICCV} or small-scale datasets with less than one hundred words~\cite{grobel1997refisolated,kulkarni2010appearance,zafrulla2011american,huang2015sign,Pigou_2017_ICCV,lim2016block,metaxas2018scalable,xue2019deep}.
These sign recognition approaches mainly consists of three steps: the feature extraction, temporal-dependency modeling and classification. 
Previous works first employ different hand-crafted features to represent static hand poses, such as SIFT-based features~\cite{yang2010chinese,yasir2015sift,tharwat2015sift}, HOG-based features~\cite{liwicki2009automatic,buehler2009learning,cooper2012sign} and features in the frequency domain~\cite{al2009video,badhe2015indian}. 
Hidden Markov Models (HMM)~\cite{starner1995visual,starner1998real} are then employed to model the temporal relationships in video sequences. Dynamic Time Warping (DTW)~\cite{lichtenauer2008sign} is also exploited to handle differences of sequence lengths and frame rates. Classification algorithms, such as Support Vector Machine (SVM)~\cite{nagarajan2013static}, are used to label the signs with the corresponding words.

Similar to action recognition, some recent works~\cite{shin2019korean,kishore2018selfie} use CNNs to extract the holistic features from image frames and then use the extracted features for classification.
Several approaches~\cite{ko2018sign,ko2019neural} first extract body keypoints and then concatenate their locations as a feature vector. The extracted features are then fed into a stacked GRU for recognizing signs. These methods demonstrate the effectiveness of using human poses in the word-level sign recognition task. Instead of encoding the spatial and temporal information separately, recent works also employ 3D CNNs~\cite{huang2015sign,ye2018recognizing} to capture spatial-temporal features together. 
However, these methods are only tested on small-scale datasets. Thus, the generalization ability of those methods remains unknown. 
Moreover, due to the lack of a standard word-level large-scale sign language dataset, the results of different methods evaluated on different small-scale datasets are not comparable and might not reflect the practical usefulness of models.


To overcome the above issues in sign recognition, we propose a large-scale word-level ASL dataset, coined WLASL database. Since our dataset consists of RGB-only videos, the algorithms trained on our dataset can be easily applied to real world cases with minimal equipment requirements. Moreover, we provide a set of baselines using state-of-the-art methods for sign recognition to facilitate the evaluation of future works.


\begin{table}[t]\caption{Comparisons of our WLASL dataset with existing ASL datasets. Column ``Mean'' indicates the average number of video samples per gloss.}
\vspace{-0.5em}
\centering
\resizebox{\linewidth}{!}{
\begin{tabular}{cccccc} \toprule
    \textbf{Datasets} & \textbf{\#Gloss} &\textbf{\#Videos} & \textbf{Mean} & \textbf{\#Signers} & \textbf{Year} \\ \midrule
    Purdue RVL-SLLL~\cite{wilbur2006purdue}  & 39 & 546   & 14  & 14 &  2006 \\
    RWTH-BOSTON-50~\cite{zahedi2005combination}  & 50 & 483   & 9.7  & 3 &  2005 \\
    Boston ASLLVD~\cite{athitsos2008american}  & 2,742  & 9,794 & 3.6 & 6 &  2008     \\ \midrule
    WLASL100~~ & 100  & 2,038  & 20.4 & 97 & 2019 \\ 
    WLASL300~~ & 300  & 5,117  & 17.1 & 109 & 2019 \\
    WLASL1000 & 1,000  & 13,168  & 13.2 & 116 & 2019 \\ 
    WLASL2000 & 2,000  & 21,083  & 10.5 & 119 & 2019 \\ 
    \bottomrule
\end{tabular}}
    \label{tab:stats}
    \vspace{-1em}
\end{table}

\section{Our Proposed WLASL Dataset}\label{sec:dataset}

In this section, we introduce our proposed Word-Level American Sign Language dataset (WLASL). We first explain the data sources and the data collection process. Following with the description of our annotation process which combines automatic detection procedures with manual annotations to ensure the correctness between signs and their annotations. Finally, we provide statistics of our WLASL.

\subsection{Dataset Collection}
In order to construct a large-scale signer-independent ASL dataset, we resort to two main sources from Internet.
First, there are multiple educational sign language websites, such as ASLU~\cite{aslu} and ASL-LEX~\cite{caselli2017asl}, and they provide lookup function for ASL signs. The mappings between glosses and signs from those websites are accurate since those videos have been checked by experts before uploaded.
Another main source is ASL tutorial videos on YouTube. We select videos whose titles clearly describe the gloss of the sign.
In total, we access 68,129 videos of 20,863 ASL glosses from 20 different websites. 
In each video, a signer performs only one sign (possibly multiple repetitions) in a nearly-frontal view with different backgrounds.

After collecting all the resources for the dataset, if the gloss annotations are composed of more than two words in English, we will remove those videos to ensure that the dataset contains words only.  
If the number of the videos for one gloss is less than seven, we also remove that gloss to guarantee that enough samples are split into the training and testing sets. Since most of the websites include daily used words, the small number of video samples for one gloss may imply those words are not frequently used. Therefore, removing those glosses with few video samples will not affect the usefulness of our dataset in practice.
After this preliminary selection procedure, we have 34,404 video samples of 3,126 glosses for further annotations.

\subsection{Annotations}
In addition to providing a gloss label for each video, some meta information, including temporal boundary, body bounding box, signer annotation and sign dialect/variation annotations, is also given in our dataset. 

\emph{Temporal boundary:} A temporal boundary is used to indicate the start and end frames of a sign. 
When the videos do not contain repetitions of signs, the boundaries are labelled as the first and last frames of the signs. Otherwise, we manually label the boundaries between the repetitions. 
For the videos containing repetitions, we only keep one sample of the repeated sign to ensure samples in which the same signer performs the same sign will not appear in both training and testing sets. Thus, we prevent learned models from overfiting to the testing set.

\emph{Body Bounding-box:} In order to reduce side-effects caused by backgrounds and let models focus on the signers, we use YOLOv3~\cite{redmon2018yolov3} as a person detection tool to identify body bounding-boxes of signers in videos.
Note that, the size of the bounding-box will change as a person signs, we use the largest bounding-box size to crop the person from the video.

\emph{Signer Diversity:} 
A good sign recognition model should be robust to inter-signer variations in the input data, \eg signer appearance and signing paces, in order to generalize well to real-world scenarios. For example, as shown in Fig.~\ref{fig:scream}, the same sign is performed with slightly different hand positioning by two signers.
From this perspective, sign datasets should have a diversity of signers. Therefore, we identify signers in our collected dataset and then provide the IDs of the signers as the meta information of the videos.
To this end, we first employ the face detector and the face embedding provided by FaceNet~\cite{schroff2015facenet} to encode faces of the dataset, and then compare the Euclidean distances among the face embeddings. If the distance between two embeddings is lower than our pre-defined threshold (\ie, 0.9), we consider those two videos signed by the same person. After automatic labeling, we also manually check the identification results and correct the mislabelled ones.

\emph{Dialect Variation Annotation:} Similar to natural languages, ASL signs also have dialect variations~\cite{mccaskill2011hidden} and those variations may contain 
different sign primitives, such as hand-shapes and motions.
To avoid the situation where dialect variations only appear in testing dataset, we manually label the variations for each gloss.
%
%
Our annotators receive training in advance to ensure that they understand the basic knowledge of ASL, in order to distinguish the differences from the signers variations and dialect variations.
To speed up the annotation process and control the annotation quality, we design an interface which lets the annotators only compare signs from two videos displayed simultaneously. Then we count the number of dialects and assign labels for different dialects automatically. 
After the dialect annotation, we also give each video a dialect label. With the help of the dialect labels, we can guarantee the dialect signs in the testing set have corresponding training samples.
We also discard the sign variations with less than five examples since there are not enough samples to be split into training, validation and testing sets. Furthermore, we notice that these variations are usually not commonly used in daily life.
%

%

\subsection{Dataset Arrangement}

After obtaining all the annotations for each video, we obtain videos with lengths ranging from 0.36 to 8.12 seconds, and the average length of all the videos is 2.41 seconds. 
The average intra-class standard deviation of the videos is 0.85 seconds. 

We sort the glosses in a descending order in terms of the sample number of a gloss. 
To provide better understanding on the difficulties of the word-level sign recognition task and the scalability of sign recognition methods, we conduct experiments on the datasets with different vocabulary sizes. In particular, we select top-$K$ glosses with $K=\{100,\,300,\,1000,\,2000\}$, and organize them to four subsets, named WLASL100, WLASL300, WLASL1000 and WLASL2000, respectively.

In Table~\ref{tab:stats}, we present statistics of the four subsets of WLASL. As indicated by Table~\ref{tab:stats}, we acquire 21,083 video samples with a duration of around 14 hours for WLASL2000 in total, and each gloss in WLASL2000 has 10.5 samples on average, which is almost three times larger than the existing large-scale dataset Boston ASLLVD. We show example frames of our dataset in Fig.~\ref{fig:signer_profiles}.


%
%

\begin{figure*}
    \centering
    \includegraphics[width=0.85\textwidth]{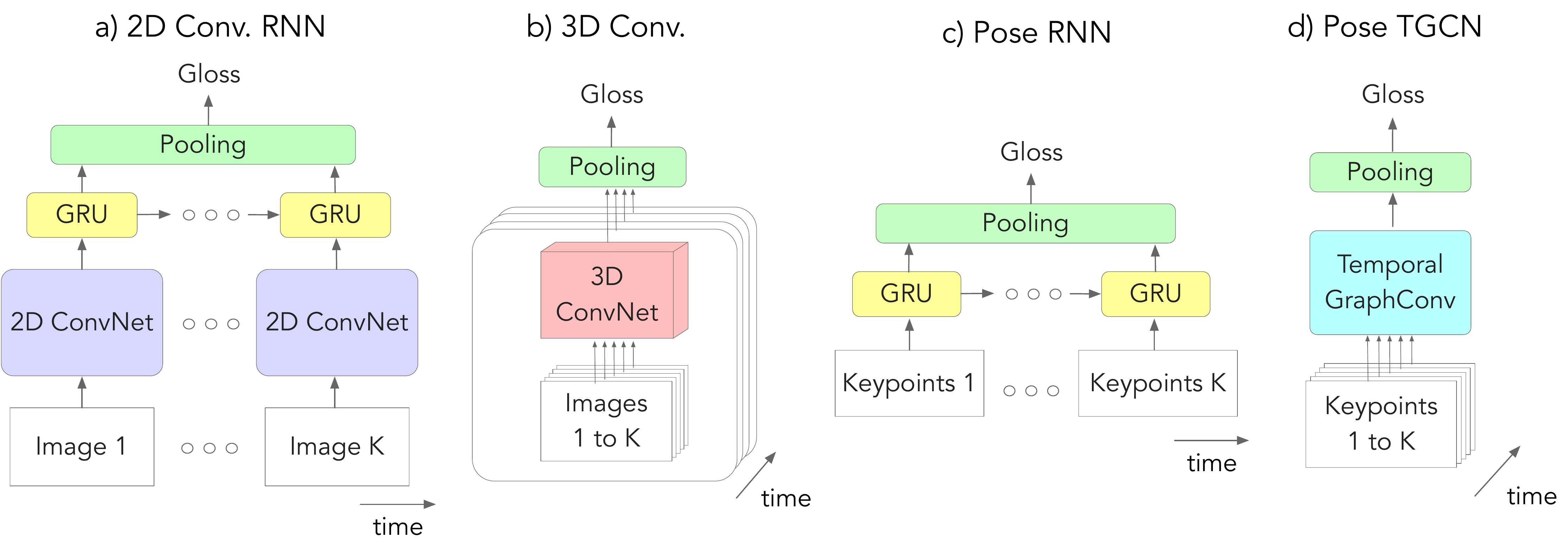}
    \vspace{-1em}
    \caption{Illustrations of our baseline architectures.}
    \label{fig:arch}
    \vspace{-1.5em}
\end{figure*}

\section{Method Comparison on WLASL}
Signing, as a part of human actions, shares similarities with human action recognition and pose estimation.
In this section, we first introduce some relevant works on action recognition and human pose estimation. Inspired by network architectures of action recognition, we employ image-appearance based and pose based baseline models for word-level sign recognition. By doing so, we not only investigate the usability of our collected dataset but also exam the sign recognition performance of deep models based on different modalities.


\subsection{Image-appearance based Baselines}
%
Early approaches employ handcrafted features to represent the spatial-temporal information from image frames and then ensemble them as a high-dimensional code for classification~\cite{laptev2008learning, wang2009evaluation, scovanner20073, laptev2005space, dalal2005histograms, tran2015learning, wang2011action}. 

Benefiting from the powerful feature extraction ability of deep neural networks, the works \cite{simonyan2014two, tran2015learning} exploit deep neural networks to generate a holistic representation for each input frame and then use the representations for recognition. 
To better establish the temporal relationship among the extracted visual features, Donahue \etal \cite{donahue2015long} and Yue \etal \cite{yue2015beyond} employ use recurrent neural networks (\eg, LSTM).  
Some works \cite{du2017rpan, cao2017realtime} also employ the joint locations as a guidance to extract local deep features around the joint regions.


Sign language recognition, especially word-level recognition, needs to focus on detailed differences between signs, such as the orientation of hands and movement direction of the arms, while the background context does not provide any clue for recognition. 
Motivated by the action recognition methods, we employ two image-based baselines to model the temporal and spatial information of videos in different manners.

\subsubsection{2D Convolution with Recurrent Neural Networks}\label{sec:vgggru}

2D Convolutional Neural Networks (CNN) are widely used to extract spatial features of input images while Recurrent Neural Networks (RNN) are employed to capture the long-term temporal dependencies among inputs. 
Thus, our first baseline is constructed by a CNN and a RNN to capture spatio-temporal features from input video frames. In particular, we use VGG16~\cite{simonyan2014very} pretrained on ImageNet to extract spatial features and then feed the extracted features to a stacked GRU~\cite{cho2014learning}. This baseline is referred to as \emph{2D Conv RNN}, and the network architecture is illustrated in Figure \ref{fig:arch}.

To avoid overfiting the training set, the hidden sizes of GRU for the four subsets are set to $64$, $96$, $128$ and $256$ respectively, and the number of the stacked recurrent layers in GRU is set to $2$.
In the training phase, we randomly select at most $50$ consecutive frames from each video. Cross-entropy losses is imposed on the output at all the time steps as well as the output feature from the average pooling of all the output features. In testing, we consider all the frames in the video and make predictions based on the average pooling of all the output features. 
%
%

\subsubsection{3D Convolutional Networks}
3D convolutional networks \cite{carreira_CVPR_2017_i3d, tran2015learning, taylor2010convolutional, ji20123d} are able to establish not only the holistic representation of each frame but also the temporal relationship between frames in a hierarchical fashion. 
Carreira \etal \cite{carreira_CVPR_2017_i3d} inflate 2D filters of the Inception network \cite{szegedy2015going} trained on ImageNet \cite{ILSVRC15}, thus obtaining well-initialized 3D filters. The inflated 3D filters are also fine-tuned on the Kinetics dataset \cite{carreira_CVPR_2017_i3d} to better capture the spatial-temporal information in a video. 



In this paper, we employ the network architecture of I3D \cite{carreira_CVPR_2017_i3d} as our second image-appearance based baseline, and the network architecture is illustrated in Figure \ref{fig:arch}. As mentioned above, the original I3D network is trained on ImageNet \cite{ILSVRC15} and fine-tuned on Kinetics-400 \cite{carreira_CVPR_2017_i3d}. In order to model the temporal and spatial information of the sign language, such as focusing on the hand shapes and orientations as well as arm movements, we need to fine-tune the pre-trained I3D. In this way, the fine-tuned I3D can better capture the spatio-temporal information of signs.
Since the class number varies in our WLASL subsets, only the last classification layer is modified in accordance with the class number.


\subsection{Pose-based Baselines}
Human pose estimation aims at localizing the keypoints or joints of human bodies from a single image or videos. 
Traditional approaches employ the probabilistic graphical model~\cite{yang2011articulated} or pictorial structures~\cite{pishchulin2013poselet} to estimate single-person poses. 
Recently, deep learning techniques have boosted the performance of pose estimation significantly. There are two mainstream approaches: regressing the keypoint positions~\cite{toshev2014human,carreira2016human}, and estimating keypoint heatmaps followed by a non-maximal suppression technique~\cite{cao2018openpose,chu2017multi,chu2016structured,yang2016end}.
However, pose estimation only provides the locations of the body keypoints, while the spatial dependencies among the estimated keypoints are not explored. 

Several works \cite{jhuang2013towards, wang2013approach} exploit human poses to recognize actions. The works \cite{jhuang2013towards, wang2013approach} represent the locations of body joints as a feature representation for recognition. These methods can obtain high recognition accuracy when the oracle annotations of the joint locations are provided. 
In order to exploit the pose information for SLR, the spatial and temporal relationships among all the keypoints require further investigation.

\subsubsection{Pose based Recurrent Neural Networks}
%
Pose based approaches mainly utilize RNNs~\cite{martinez2017human} to model the pose sequences for analyzing human motions. Inspired by this idea, our first pose-based baseline employs RNN to model the temporal sequential information of the pose movements, and the representation output by RNN is used for the sign recognition. 

In this work, we extract 55 body and hand 2D keypoints from a frame on WLASL using OpenPose~\cite{cao2018openpose}. These keypoints include 13 upper-body joints and 21 joints for both left and right hands as defined in~\cite{cao2018openpose}.
Then, we concatenate all the 2D coordinates of each joint as the input feature and feed it to a stacked GRU of 2 layers. 
In the design of GRUs, we use the empirically optimized hidden sizes of $64$, $64$, $128$ and $128$ for the four subsets respectively.
Similar to the training and testing protocols in Section~\ref{sec:vgggru}, $50$ consecutive frames are randomly chosen from the input video. Cross-entropy losses is employed for training. In testing, all the frames in a video are used for classification.

\subsubsection{Pose Based Temporal Graph Neural Networks}
We introduce a novel pose-based approach to ISLR using \emph{Temporal Graph Convolution Networks (TGCN)}. Consider the input pose sequence $\mathbf{X}_{1:N}=[\mathbf{x}_1, \mathbf{x}_2,\mathbf{x}_3,...,\mathbf{x}_N]$ in $N$ sequential frames, where $\mathbf{x_i}\in\mathbb{R}^K$ represents the concatenated 2D keypoint coordinates in dimension K. We propose a new graph network based architecture that models the spatial and temporal dependencies of the pose sequence. 
Different from existing works on human pose estimation, which usually model motions using 2D joint angles, we encode temporal motion information as a holistic representation of the trajectories of body keypoints. 

\begin{figure}
    \centering
    \includegraphics[width=0.6\linewidth]{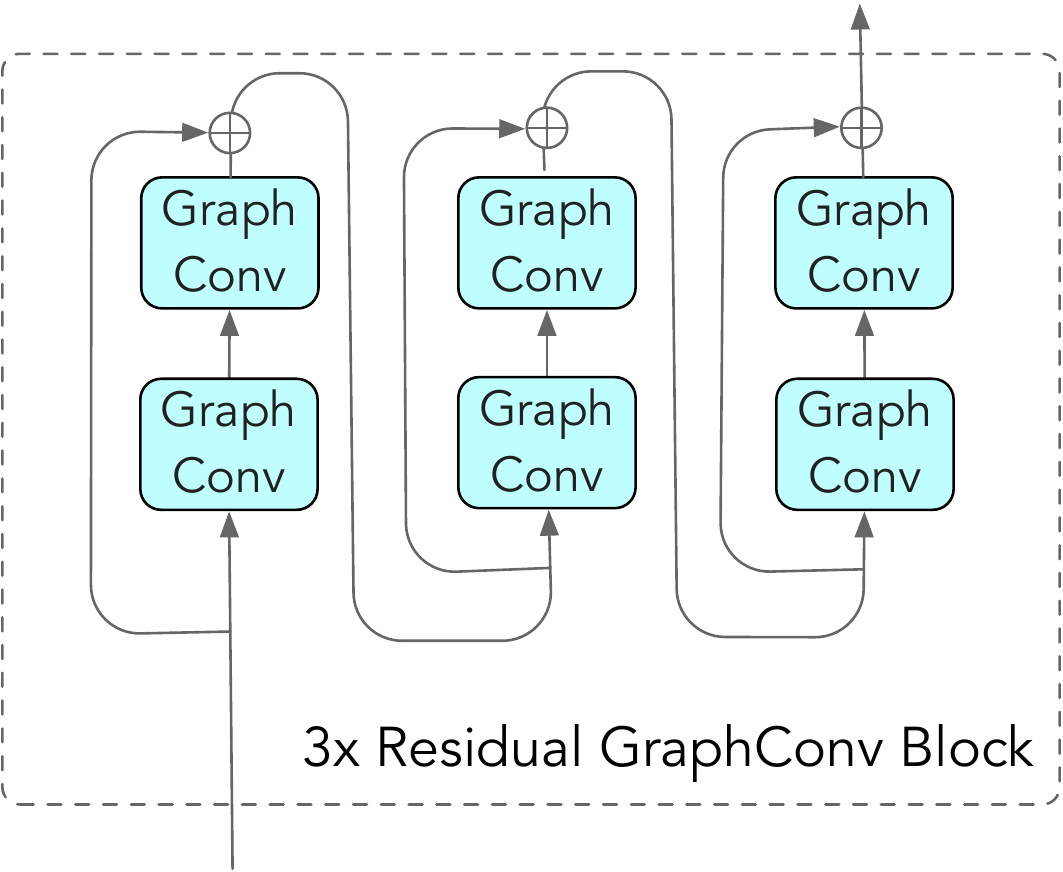}
    \vspace{-1em}
    \caption{Residual Graph Convolution Block.}
    \label{fig:resgc}
    \vspace{-2em}
\end{figure}

Motivated by the recent work on human pose forecasting \cite{chiu2019action, chiu2019action}, we view a human body as a fully-connected graph with $K$ vertices and represent the edges in the graph as a weighted adjacency matrix $\mathbf{A}\in\mathbb{R}^{K\times K}$.
Although a human body is only partially connected, we construct the human body as fully-connected graph in order to learn the dependencies among joints via a graph network.
In a deep graph convolutional network, the n-th graph layer is a function $\mathcal{G}_{n}$ that takes as input features a matrix $\mathbf{H}_{n}\in\mathbb{R}^{K\times F}$, where $F$ is the feature dimension output by its previous layer.
In the first layer, the networks takes as input the $K\times 2N$ matrix coordinates of body keypoints. Given this formulation and a set of trainable weights $\mathbf{W}_n\in \mathbb{R}^{F\times F^\prime}$, a graph convolutional layer is expressed as:
\vspace{-0.7em}
\begin{align}
    \mathbf{H}_{n+1} = \mathcal{G}_n(\mathbf{H}_{n})= \sigma(\mathbf{A}_n\mathbf{H}_n\mathbf{W}_n),
\end{align}
where $\mathbf{A}_n$ is a trainable adjacency matrix for n-th layer and $\sigma(\cdot)$ denotes the activation function $tanh(\cdot)$. A residual graph convolutional block stacks two graph convolutional layers with a residual connection as shown in Fig.~\ref{fig:resgc}.
Our proposed TGCN stacks multiple residual graph convolutional blocks and takes the average pooling result along the temporal dimension as the feature representation of pose trajectories. Then a softmax layer followed by the average pooling layer is employed for classification.

\subsection{Training and Testing Protocol}
\subsubsection{Data Pre-processing and Augmentation}
We resize the resolution of all original video frames such that the diagonal size of the person bounding-box is $256$ pixels. 
For training VGG-GRU and I3D, we randomly crop a $224\times224$ patch from an input frame and apply a horizontal flipping with a probability of 0.5. Note that, the same crop and flipping operations are applied to the entire video frames instead of in a frame-wise manner. Similar to \cite{carreira2017quo}, when training VGG-GRU, Pose-GRU and Pose-TGCN, for each video consecutive 50 frames are randomly selected and the models are asked to predict labels based on only partial observations of the input video. In doing so, we increase the discriminativeness of the learned model. For I3D, we follow its original training configuration. 

\subsubsection{Implementation details}
The models, \ie, VGG-GRU, Pose-GRU, Pose-TGCN and I3D are implemented in PyTorch. It is important to notice that we use the I3D pre-train weights provided by Carreira \etal \cite{carreira_CVPR_2017_i3d}.
We train all the models with Adam optimizer~\cite{kingma2014adam}. Note that, I3D was trained by stochastic gradient descent (SGD) in~\cite{carreira2017quo}. However, I3D does not converge when using SGD to fine-tune it in our experiments. Thus, Adam is employed to fine-tune I3D.
%
All the models are trained with 200 epochs on each subset. We terminate the training process when the validation accuracy stops increasing. 

We split the samples of a gloss into the training, validation and testing sets following a ratio of 4:1:1. We also ensure each split has at least one sample per gloss. The split information will be released publicly as part of WLASL.

\subsubsection{Evaluation Metric}
We evaluate the models using the mean scores of top-$K$ classification accuracy with $K=\{1,\,5,\,10\}$ over all the sign instances. 
As seen in Figure \ref{fig:rice-soup}, different meanings have very similar sign gestures, and those gestures may cause errors in the classification results. However, some of the erroneous classification can be rectified by contextual information. Therefore, it is more reasonable to use top-K predicted labels for the word-level sign language recognition.  
%

\subsection{Discussion}
\begin{table*}[t]\caption{Top-1, top-5, top-10 accuracy ($\%$) achieved by each model (by row) on the four WLASL subsets.}
\vspace{-0.5em}
\centering
\resizebox{\linewidth}{!}{
\begin{tabular}{cccc|ccc|ccc|ccc} \toprule
     {Method} & \multicolumn{3}{c}{{WLASL100}} & \multicolumn{3}{c}{{WLASL300}} & \multicolumn{3}{c}{{WLASL1000}} & \multicolumn{3}{c}{{WLASL2000}} \\ 
      & top-1 & top-5 & top-10 & top-1& top-5& top-10& top-1& top-5& top-10& top-1 & top-5 & top-10\\ \midrule
    Pose-GRU  & 46.51 & 76.74 & 85.66 & 33.68 & 64.37 & 76.05 & 30.01 & 58.42 & 70.15 & 22.54 & 49.81 & 61.38 \\
    Pose-TGCN  & 55.43 & 78.68 & 87.60 & 38.32 & 67.51 & 79.64 & 34.86 & 61.73 & 71.91 & 23.65 & 51.75 & 62.24 \\ \midrule
    VGG-GRU  & 25.97 & 55.04 & 63.95 & 19.31 & 46.56 & 61.08 & 14.66 & 37.31 & 49.36 & 8.44 & 23.58 & 32.58\\
    I3D  & \textbf{65.89} & \textbf{84.11} & \textbf{89.92} & \textbf{56.14} & \textbf{79.94} & \textbf{86.98} & \textbf{47.33} & \textbf{76.44} & \textbf{84.33} & \textbf{32.48} & \textbf{57.31} & \textbf{66.31}  \\
    \bottomrule
\end{tabular}}
    \label{tab:exp-results}
    \vspace{-0em}
\end{table*}
\begin{table*}[t]\caption{Top-10 accuracy ($\%$) of I3D (and Pose-TGCN when trained (row) and tested (column) on different WLASL subsets.}
\vspace{-0.5em}
\centering
\begin{tabular}{lcc|cc|cc|cc} \toprule
      &
  \multicolumn{2}{c}{{WLASL100}} & \multicolumn{2}{c}{{WLASL300}} & \multicolumn{2}{c}{{WLASL1000}} & \multicolumn{2}{c}{{WLASL2000}} \\
      & I3D & TGCN & I3D & TGCN & I3D & TGCN & I3D & TGCN \\
      \midrule
     WLASL100  & 89.92  & 87.60 & - & - & -  & - & - & - \\
     WLASL300  & 88.37 & 81.40 & 86.98 & 79.64 & - & - & - & - \\ 
     WLASL1000 & 85.27 & 77.52 & 86.22 & 74.25 & 84.33 & 71.91 & - & - \\
     WLASL2000 & 72.09 & 67.83 & 71.11 & 65.42 & 67.32 & 64.55 & 66.31 & 62.24\\
    \bottomrule
\end{tabular}
    \label{tab:eff-num-cls}
    \vspace{-0.5em}
\end{table*}

\subsubsection{Performance Evaluation of Baseline Networks}
%


Table~\ref{tab:exp-results} indicates that the performance of our baseline models based on poses and image-appearance. The results demonstrate that our pose-based TGCN further improves the classification accuracy in comparison to the pose-based sign recognition method Pose-GRU. This indicates that our proposed pose-TGCN captures both spatial and temporal relationships of the body keypoints since Pose-GRU mainly explores the temporal dependencies of the keypoints for classification.
On the other hand, our fine-tuned I3D model achieves better performance compared to the other image-appearance based model VGG-GRU since I3D has larger network capacity and is pretrained on not only ImageNet but also Kinetics. 

Although I3D is larger than our TGCN, Pose-TGCN can still achieve comparable results with I3D at top-5 and top-10 accuracy on the large-scale subset WLASL2000. This demonstrates that our TGCN effectively encodes human motion information. Since we use an off-the-shelf pose estimator \cite{cao2018openpose}, the erroneous estimation of poses may degrade the recognition performance. In contrast, image appearance-based baselines are trained in an end-to-end fashion for sign recognition and thus the errors residing in spatial features can be reduced during training. Therefore, training pose-based baselines in an end-to-end fashion could further improve the recognition performance. 

\subsubsection{Effect of Vocabulary Size}
As seen in Table~\ref{tab:exp-results}, our baseline methods can achieve relatively high classification accuracy on small-size subsets. \ie, WLASL100 and WLASL300. However, the subset WLASL2000 is very close to the real-world word-level classification scenario due to its large vocabulary. Pose-GRU, pose-TGCN and I3D achieve similar performance on WLASL2000. This implies that the recognition performance on small vocabulary datasets does not reflect the model performance on large vocabulary datasets, and the large-scale sign language recognition is very challenging. 

We also evaluate how the class number, \ie, vocabulary size, impacts on the model performance. There are two factors mainly affecting the performance: 
(i) deep models themselves favor simple and easy tasks, and thus they perform better on smaller datasets. As indicated in Table~\ref{tab:exp-results}, the models trained on smaller vocabulary size sets perform better than larger ones (comparing along columns);
(ii) the dataset itself has ambiguity. Some signs, as shown in Figure \ref{fig:rice-soup}, are hard to recognize by even humans, and thus deep models will be also misled by those classes. As the number of classes increases, there will be more ambiguous signs. 

In order to explain the impacts of the second factor, we dissect the models, \ie, I3D and Pose-TGCN, trained on WLASL2000. Here, we test our models on the WLASL100, WLASL300, WLASL1000 and WLASL2000. As seen in Table 4, when the test class number is smaller, the models achieve higher accuracy (comparing along rows). The experiments imply that as the number of classes decreases, the number of ambiguous signs becomes smaller, thus making classification easier.




%

\subsubsection{Effect of Sample Numbers}

As the class number in the dataset increases, training a deep model requires more samples. However, as illustrated in Table 1, although in our dataset each gloss contains more samples than other datasets, the number of training examples per class is still relatively small compared to some large-scale generic activity recognition datasets \cite{caba2015activitynet}. This brings some difficulties for the network training. Note that, the average training samples for each gloss in WLASL100 are twice large as those in WLASL2000. Therefore, models obtain better classification performance on the glosses with more samples, as indicated in Table 3 and Table 4.
Crowdsourcing via Amazon Mechanism Tucker (AMT) is a popular way to collect data. However, annotating ASL requires specific domian knowledge and makes crowdsourcing infeasible.



\section{Conclusion}





In this paper, we proposed a large-scale Word-Level ASL (WLASL) dataset covering a wide range of daily words and evaluated the performance of deep learning based methods on it. To the best of our knowledge, our dataset is the largest public ASL dataset in terms of the vocabulary size and the number of samples for each class. 
Since understanding sign language requires very specific domain knowledge, labelling a large amount of samples per class is unaffordable. 
After comparisons among deep sign recognition models on WLASL, we conclude that developing word-level sign language recognition algorithms on such a large-scale dataset requires more advanced learning algorithms, such as few-shot learning. In our future work, we also aim at utilizing word-level annotations to facilitate sentence-level and story-level machine sign translations. 

\section*{Acknowledgement}
This research is supported in part by the Australia Research Council ARC Centre of Excellence for Robotics Vision (CE140100016),  ARC-Discovery (DP 190102261) and ARC-LIEF (190100080). The authors gratefully acknowledge the GPU gift donated by NVIDIA Corporation.  We thank all anonymous reviewers for their constructive comments.

{\small
\bibliographystyle{ieee}
\bibliography{egbib}
}

\end{document}